# Boosting Adversarial Transferability via High-Frequency Augmentation and Hierarchical-Gradient Fusion

Yayin Zheng, Chen Wan(✉), Zihong Guo, Hailing Kuang, Xiaohai Lu

Department of Computer Science and Technology, Shantou University, Shantou, China
wanchen18@outlook.com

**Abstract.** Adversarial attacks have become a significant challenge in the security of machine learning models, particularly in the context of black-box defense strategies. Existing methods for enhancing adversarial transferability primarily focus on the spatial domain. This paper presents Frequency-Space Attack (FSA), a new adversarial attack framework that effectively integrates frequency-domain and spatial-domain transformations. FSA combines two key techniques: (1) High-Frequency Augmentation, which applies Fourier transform with frequency-selective amplification to diversify inputs and emphasize the critical role of high-frequency components in adversarial attacks, and (2) Hierarchical-Gradient Fusion, which merges multi-scale gradient decomposition and fusion to capture both global structures and fine-grained details, resulting in smoother perturbations. Our experiment demonstrates that FSA consistently outperforms state-of-the-art methods across various black-box models. Notably, our proposed FSA achieves an average attack success rate increase of 23.6% compared with BSR (CVPR 2024) on eight black-box defense models.

**Keywords:** Adversarial Attack, Transferability, Frequency-Space Attack.

## 1 Introduction

Deep neural networks (DNNs) have achieved remarkable success across various domains. However, their vulnerability to imperceptible adversarial examples (AEs) remains a critical challenge [1].These adversarial examples are crafted by introducing subtle, human-imperceptible perturbations to the original images, which can cause DNNs to produce incorrect predictions. Furthermore, AEs exhibit a concerning property known as transferability [2], where examples designed for a given model (that is, the white-box model) can also mislead other models. Investigating the transferability of AEs not only reveals the vulnerabilities of DNNs, but also provides a foundation for developing more robust defenses.

To enhance the transferability of adversarial examples, numerous methods have been proposed, with input transformation-based attacks emerging as the most widely adopted strategies. These methods apply a series of operations, such as random resizing and padding [3], translation [4], and scaling [5], to the input examples during gradient computation, aiming to prevent the generated adversarial examples from overfitting to the model. More recently, more complex input transformations have been introduced. For instance, *Admix* [6] blends input image with images of different categories, SIA [7] applies transformations to individual blocks, and BSR [8] employs block shuffling and rotation to further boost transferability. In parallel, several studies [9-15] have explored the vulnerabilities of DNNs from a frequency-domain perspective. Long et al. [13] propose SSA, a frequency-domain attack using spectral perturbations in the DCT domain with Gaussian noise and uniform spectral masks.

Despite these advances, the exploration of frequency-domain transformations remains limited. Notably, some studies have revealed two critical insights: first, DNNs exhibit distinct sensitivities to different frequency components [16]; and second, the adversarially trained models (i.e., defense models) show increased robustness when processing low-frequency information [17]. These observations suggest that high-frequency components may be more influential in crafting effective adversarial perturbations, whereas low-frequency components contribute more to enhancing model robustness. Building on these insights, we propose a new Frequency-Space Attack (FSA), which leverages the complementary strengths of both frequency-domain and spatial-domain manipulations to generate adversarial examples. The proposed FSA consists of two modules: High-Frequency Augmentation Module (HAM) in the frequency domain and Hierarchical-Gradient Fusion Module (HFM) in the spatial domain. HAM transforms input images into the frequency domain via the Fourier trans-form and amplifies their high-frequency components, while HFM operates in the spatial domain, employing a multi-scale approach that transforms gradient information into different scales and fuses them. By integrating frequency-domain feature refinement with spatial-domain multi-scale optimization, FSA establishes a unified framework for generating adversarial examples that achieve high attack success rates against the black-box defense models.

The main contributions of this paper are summarized as follows:

1) We propose a new FSA that leverages the complementary strengths of both frequency-domain and spatial-domain to enhance adversarial transferability.

2) The proposed HAM utilizes Fourier transforms and selective amplification to enhance the diversity of input examples, revealing the critical role of high-frequency information in adversarial attacks.

3) The proposed HFM employs multi-scale gradient decomposition and fusion to capture both global structures and local details, effectively mitigating the overfitting of adversarial examples to white-box models.

4) Extensive experiments demonstrate that FSA consistently outperforms state-of-the-art attack methods, achieving an average improvement of 23.6% in attack success rate against defense models compared to BSR.

## 2      Related Works

To improve the transferability of AEs, the two most commonly adopted strategies are gradient-based attacks and input transformation-based attacks. Gradient-based attacks refine gradient estimation to generate more imperceptible yet effective perturbations.

MI-FGSM [18] and NI-FGSM [5] extend I-FGSM [19] by incorporating momentum and Nesterov acceleration respectively, while PI-FGSM [20] enhances gradient updates by employing an amplification factor and redistributing cut noise through a project kernel. In contrast, input transformation-based attacks improve generalization by aggregating gradients computed over various transformed inputs. DIM [3] applies random resizing and padding, TIM [4] employs convolutional smoothing to enhance translation invariance, and SIM [5] utilizes multi-scale resampling. *Admix* [6] blends images to increase perturbation diversity, SIA [7] applies transformations to individual blocks, while BSR [8] enhances transferability by shuffling and rotating image blocks.

To explore the frequency-domain behavior of DNN, Wang et al. [14] and Yin et al. [15] found that CNNs rely heavily on high-frequency components, improving accuracy but reducing adversarial robustness. Yin et al. [15] further observed that traditional data augmentation techniques, such as Gaussian noise injection, induce a robustness trade-off between high-frequency and low-frequency perturbations. Long et al. [13] proposed SSA, a frequency-domain attack that enhances adversarial transferability by introducing spectral perturbations in the DCT domain through Gaussian noise and uniform spectral mask.

## 3    Methodology

### 3.1    Framework Overview

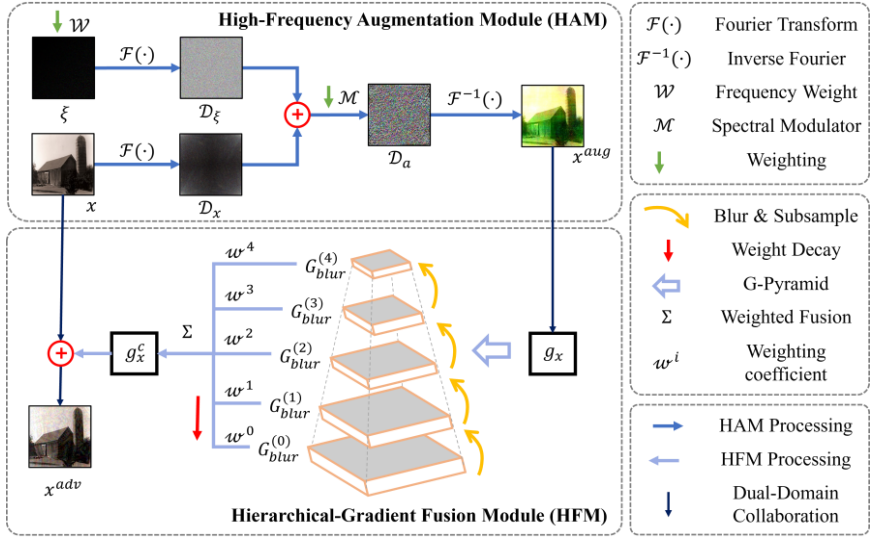

**Fig. 1.** The flowchart of our proposed FSA.

Our proposed Frequency-Space Attack (FSA) framework is illustrated in Fig. 1, which integrates two modules: the High-Frequency Augmentation Module (HAM) and the Hierarchical-Gradient Fusion Module (HFM). HAM leverages Fourier transforms and selective amplification to enhance the diversity of input examples $x$, generating augmented examples $x^{aug}$. In the subsequent HFM stage, the gradient of the augmented example $g_x$, undergoes multi-scale gradient decomposition and fusion, resulting in the combined gradient $g_x^c$, which is then utilized to determine the adversarial perturbation to craft adversarial examples $x^{adv}$.

### 3.2    High-Frequency Augmentation Module

As illustrated in Fig. 1, the proposed FSA begins with the HAM process, which generates the augmented example $x^{aug}$. In HAM, the input example $x$ with dimensions $H \times W \times 3$ is first transformed into the frequency domain using the Fourier transform $\mathcal{F}(\cdot)$, resulting in its frequency representation $\mathcal{D}_x$. Simultaneously, Gaussian noise $\xi$ is generated and mapped to the frequency domain to produce $\mathcal{D}_\xi = \mathcal{F}(\xi \cdot \mathcal{W})$. Here, each element of $\xi$ follows $\xi_{h,w} \sim \mathcal{N}(0, \sigma^2)$, while the high-frequency weighting matrix $\mathcal{W}$ is defined as $\mathcal{W}_{h,w} = (h + w)/(H + W - 2)$, where $h$ and $w$ represent the coordinate positions in $x$. This weighting function increases with spatial frequency, amplifying high-frequency components.

After obtaining $\mathcal{D}_x$ and $\mathcal{D}_\xi$, the high-frequency enhanced noise $\mathcal{D}_\xi$ is integrated into $\mathcal{D}_x$, followed by the introduction of a random spectral modulation matrix $\mathcal{M}$ for adjustment, resulting in $\mathcal{D}_a$, formulated as:

$$\mathcal{D}_a = \left(\mathcal{D}_x + \mathcal{D}_\xi\right) \cdot \mathcal{M} \tag{1}$$

where $\mathcal{M} \sim U(1 - \rho, 1 + \rho)$ is a random frequency-domain mask, and $ho$ modulates the intensity of frequency-domain adjustments, and the weighting matrix $\mathcal{W}$ dynamically regulates $\rho$. The augmented example $x^{aug}$ is then reconstructed through inverse Fourier transformation:

$$x^{aug} = \mathcal{F}^{-1}(\mathcal{D}_a) \tag{2}$$

where $\mathcal{F}^{-1}(\cdot)$ denotes the Inverse Fourier Transform.

### 3.3    Hierarchical-Gradient Fusion Module

By incorporating the aforementioned FSA, we obtain the augmented example $x^{aug}$. The gradient of the loss function with respect to $x^{aug}$, denoted as $g_x$, is then computed to guide the adversarial perturbation.

In the HFM of Fig. 1, a Gaussian pyramid is constructed by applying Gaussian smoothing and down-sampling to $g_x$, producing a series of progressively lower-resolution gradient representations, namely $G_{blur}^{(0)}$, $G_{blur}^{(1)}$, ..., $G_{blur}^{(n-1)}$, where $n$ denotes the number of layers in the pyramid. The mathematical form of $G_{blur}^{(i)}$ is given by:

$$G_{blur}^{(0)} = g_x \tag{3}$$

$$G_{blur}^{(i)} = F_{down}\left(\mathcal{G}\left(G_{blur}^{(i-1)}\right)\right) \tag{4}$$

where $F_{down}(\cdot)$ represents the down-sampling operation, and $\mathcal{G}(\cdot)$ denotes the Gaussian blur operation. The index $i \in [1, n-1]$ denotes the pyramid level, where each $G_{blur}^{(i)}$ is recursively generated from the previous level $G_{blur}^{(i-1)}$ via Gaussian blur and down-sampling.

To aggregate information across different scales, we fuse the pyramid layers using a weighted summation:

$$g_x^c = \sum_{i=0}^{L-1} w_i \cdot F_{up}(G_{blur}^{(i)}) \tag{5}$$

where $g_x^c$ denotes the combined gradient, $F_{up}(\cdot)$ up-samples each level to the same dimension as $g_x$, and the weighting coefficient $w_i$ is defined as:

$$w_i = \frac{\beta^{L-i}}{\sum_{k=0}^{L} \beta^{L-k}} \tag{6}$$

where decay factor $\beta$ lies within the interval (0,1), ensuring that lower-resolution gradients contribute more significantly to the combined gradient $g_x^c$. This strategy integrates coarse-grained structural information from deeper pyramid layers while preserving fine-grained details from shallower layers. The multi-scale fusion mechanism adaptively balances local and global gradient information during optimization, ensuring more effective gradient-guided updates.

### 3.4 Frequency-Space Attack

According to equation (5), we derive the combined gradient $g_x^c$, which forms the core of our algorithm. In an iterative attack, each step uses the current adversarial example $x_t^{adv}$ (where $x_0^{adv} = x$) as input to compute the corresponding combined gradient $g_{x_t}^c$. This gradient then guides the update of the adversarial perturbation for $x_t^{adv}$. By incorporating $g_{x_t}^c$ into MI-FGSM, we develop the FSA method. The process of generating adversarial examples using FSA is summarized as follows:

$$g_{t+1} = \mu \cdot g_t + \frac{g_{x_t}^c}{\|g_{x_t}^c\|_1} \tag{7}$$

$$x_{t+1}^{adv} = Clip_x^\epsilon \{x_t^{adv} + \alpha \cdot sign(g_{t+1})\} \tag{8}$$

where $\mu$ denotes the momentum factor, $\epsilon$ denotes the maximum perturbation, $\alpha$ denotes the step size, $\alpha = \epsilon/T$, $T$ represents the number of iterations, $sign(\cdot)$ denotes the sign function, and $Clip_x^\epsilon\{\cdot\}$ ensures that the generated adversarial example is clipped within the $\epsilon$-ball of the original image $x$.

## 4    Experiment

### 4.1    Settings

We implement our framework in TensorFlow and perform evaluations using an NVIDIA RTX 4090 Ti GPU. To ensure comparability with previous studies, we adopt the $l$-norm as the metric for distortion and utilize the cross-entropy loss during training.

**Dataset**: The test dataset comprises 1000 randomly sampled images from the ImageNet validation set [21], as provided by Lin et al [5].

**Models**: We select four normally trained models—Inc-v3 [22], Inc-v4 [23], IncRes-v2 [23], and Res-v2-101 [24] as white-box models to generate AEs. The adversarial examples are evaluated on eight defense models, including Inc-v3$_{ens3}$ [25], Inc-v3$_{ens4}$ [25], IncRes-v2$_{ens}$ [25], HGD [26], R&P [27], NIPS-r3[1], FD [28], and NRP [29].

**Baselines**: Six input transformation-based attacks are selected as our baselines, including DIM [3], TIM [4], SIM [5], *Admix* [6], SSA [13] and BSR [8]. For fairness, all input transformations are integrated into MI-FGSM [18]. In addition, we compare the combination versions of these methods, such as STD (a combination of MI-FGSM [18], SIM [5], TIM [4], and DIM [3]) and FSA-STD.

**Hyper-parameters**: The maximum perturbation $\epsilon = 16$, number of iterations $T = 10$, and step size $\alpha = \epsilon/T = 1.6$. MI-FGSM uses a momentum factor $\mu = 1.0$. DIM has a transformation probability $p = 0.5$. TIM employs a $7 \times 7$ kernel. SIM and *Admix* use $m = 5$ copies (with *Admix* using $m_2 = 3$ examples and a mixing ratio $\eta = 0.2$). For SSA, we set the tuning factor $\rho = 0.5$, the standard deviation $\sigma$ of $\xi$ to $\epsilon$, and the number of spectral transformations $N = 20$. BSR divides images into $2 \times 2$ blocks, and applies $\tau = 24°$ and $N = 20$ transformations. For our proposed FSA, we set decay factor $\beta = 0.8$, pyramid layers $n = 5$, spectral factor $\rho = 0.7$, the standard deviation $\sigma$ of $\xi$ to $2\epsilon$, and the number of frequency-space attacks to $N = 20$. It is worth noting that in Sec. 4.3: Evaluation on Combined Input Transformation, to reduce spatial complexity and balance transformation intensity, we set $N = 8$. The other parameter settings remain the same as in the individual attack methods.

### 4.2    Evaluation on Single Model

This section analyzes the ASR of adversarial examples generated by FSA on four standard trained models (Inc-v3, Inc-v4, IncRes-v2, Res-101) against eight defense models and compares it with other mainstream adversarial attack methods (DIM, TIM, SIM, *Admix*, SSA, and BSR). The values in the Table 1 represent the ASR, i.e., the misclassification rate of the target model. Each column represents the attacked model, while each row represents adversarial examples generated by the corresponding source model. The results indicate that FSA achieves a significantly higher average ASR on the source models compared to other methods, with an improvement ranging from 2.8% to 46.5%. Furthermore, the average ASR of FSA exceeds that of BSR by 2.8% to 19.1%. Specifically, compared with traditional methods such as DIM, TIM, and SIM,

---

FSA improves the average success rate by 25.9% to 51.0%, further verifying its effectiveness in enhancing adversarial transferability. On the IncRes-v2$_{ens}$ ensemble model, when using Inc-v3 as the source model, FSA achieves an attack success rate of 64.1%, significantly outperforming BSR and SSA. Additionally, across other source models, FSA outperforms BSR and SSA by 13.9% to 29.6%. These results demonstrate the superiority of FSA in generating transferable adversarial examples and further emphasize the importance of frequency-space collaborative attacks as a means to enhance transferability.

**Table 1.** ASR (%) of eight models under different input transformations in a single-model setting.

| Model | Attack | Inc-v3$_{ens3}$ | Inc-v3$_{ens4}$ | IncRes-v2$_{ens}$ | HGD | R&P | NIPS-r3 | NRP | FD | AVG. |
|-------|--------|---------|---------|-----------|------|------|---------|------|------|------|
| | DIM | 18.5 | 17.6 | 9.4 | 6.9 | 8.0 | 14.2 | 16.1 | 45.7 | 17.1 |
| | TIM | 24.1 | 21.0 | 12.8 | 16.9 | 12.0 | 15.1 | 17.3 | 42.1 | 20.2 |
| | SIM | 32.6 | 31.1 | 17.6 | 14.7 | 16.4 | 23.7 | 24.3 | 54.7 | 26.9 |
| Inc-v3 | *Admix* | 40.1 | 39.0 | 21.1 | 21.4 | 20.6 | 28.9 | 28.1 | 63.5 | 32.8 |
| | SSA | 40.6 | 38.2 | 23.3 | 14.5 | 23.0 | 30.2 | 33.9 | 69.0 | 34.1 |
| | BSR | 53.8 | 50.6 | 30.8 | 42.3 | 33.0 | 43.8 | 28.8 | 72.9 | 44.5 |
| | FSA | **74.8** | **72.7** | **64.1** | **66.1** | **59.9** | **66.2** | **64.3** | **76.5** | **68.1** |
| | DIM | 22.1 | 20.8 | 10.1 | 10.6 | 12.3 | 16.7 | 15.5 | 45.8 | 19.2 |
| | TIM | 25.9 | 23.9 | 17.5 | 22.5 | 17.1 | 19.4 | 15.9 | 43.8 | 23.3 |
| | SIM | 47.5 | 45.1 | 29.3 | 29.4 | 29.2 | 36.6 | 29.8 | 59.2 | 38.3 |
| Inc-v4 | *Admix* | 54.9 | 51.4 | 33.0 | 35.9 | 33.8 | 42.6 | 33.9 | 68.0 | 44.2 |
| | SSA | 46.1 | 43.8 | 31.9 | 25.9 | 29.2 | 38.1 | 36.3 | 68.3 | 40.0 |
| | BSR | 56.0 | 52.3 | 34.3 | 52.3 | 38.1 | 48.2 | 30.0 | 72.1 | 47.9 |
| | FSA | **69.4** | **70.4** | **63.9** | **53.6** | **61.1** | **65.3** | **56.3** | **73.0** | **64.1** |
| | DIM | 32.2 | 25.1 | 17.7 | 19.2 | 18.3 | 23.8 | 19.2 | 48.7 | 25.5 |
| | TIM | 32.2 | 27.7 | 21.9 | 26.2 | 21.0 | 24.0 | 19.7 | 50.0 | 27.8 |
| | SIM | 57.1 | 49.1 | 41.7 | 40.0 | 35.8 | 43.1 | 34.5 | 64.9 | 45.8 |
| IncRes-v2 | *Admix* | 61.1 | 52.1 | 45.7 | 43.5 | 40.1 | 48.6 | 39.4 | 70.5 | 50.1 |
| | SSA | 56.5 | 52.1 | 46.1 | 42.5 | 42.3 | 49.5 | 42.4 | 70.3 | 50.2 |
| | BSR | 68.7 | 62.0 | 49.7 | 64.7 | 54.8 | 63.0 | 37.5 | **77.1** | 59.7 |
| | FSA | **74.7** | **72.5** | **73.4** | **72.5** | **69.4** | **71.3** | **66.5** | 73.5 | **71.7** |
| | DIM | 36.1 | 32.7 | 20.9 | 27.8 | 22.4 | 30.3 | 24.4 | 56.5 | 31.4 |
| | TIM | 36.4 | 32.3 | 22.9 | 31.4 | 24.1 | 28.2 | 24.3 | 51.5 | 31.4 |
| | SIM | 43.4 | 38.3 | 26.3 | 32.3 | 25.8 | 33.0 | 29.0 | 59.0 | 35.9 |
| Res-101 | *Admix* | 48.7 | 42.1 | 30.5 | 35.8 | 29.0 | 37.9 | 32.3 | 64.8 | 40.1 |
| | SSA | 53.0 | 50.2 | 39.1 | 44.3 | 37.9 | 47.3 | 41.7 | 74.1 | 48.5 |
| | BSR | **78.8** | 73.2 | 54.9 | **79.3** | 61.3 | **72.9** | 45.0 | **85.5** | 68.9 |
| | FSA | 76.4 | **73.7** | **68.8** | 70.4 | **67.0** | 70.5 | **68.0** | 79.0 | **71.7** |

### 4.3   Evaluation on Combined Input Transformation

Previous studies have shown that a well-designed input transformation-based attack not only exhibits better transferability but can also be compatible with other input transformations, thereby generating more transferable adversarial examples. To evaluate this, we integrate our proposed FSA with various input transformations, denoted as DIM-FSA, TIM-FSA, SIM-FSA, STD-FSA, *Admix*-FSA, and BSR-FSA. We report the attack success rates of adversarial examples generated on Inc-v3 in Table 2. Our results demonstrate that FSA significantly enhances the transferability of these input transformation-based attacks. When FSA is combined with these transformations, the ASR improves by 18.5% to 49.2% on average. Notably, when FSA is integrated with SI-DI-TIM, it further enhances transferability by 7.8% to 39.1%. Additionally, combining FSA with BSR remarkably boosts the attack success rate by an impressive 41.5% on average. These findings highlight the strong compatibility of FSA with existing input transformation strategies, demonstrating that its combination with other techniques can further optimize attack generalization and enhance adversarial transferability.

**Table 2.** ASR (%) of eight models under different transformations combined with FSA. ↑ indicates ASR improvement.

| Attack | Inc-v3$_{ens3}$ | Inc-v3$_{ens4}$ | IncRes-v2$_{ens}$ | HGD | R&P | NIPS-r3 | NRP | FD | AVG.↑ |
|---|---|---|---|---|---|---|---|---|---|
| DIM-FSA | 73.2$_{+54.7}$ | 70.7$_{+53.3}$ | 62.1$_{+52.7}$ | 58.2$_{+51.3}$ | 61.2$_{+53.2}$ | 65.6$_{+51.4}$ | 63.1$_{+47.0}$ | 76.0$_{+30.3}$ | 49.2 |
| TIM-FSA | 68.6$_{+44.5}$ | 67.0$_{+46.0}$ | 57.2$_{+44.4}$ | 64.2$_{+47.3}$ | 54.3$_{+42.3}$ | 58.4$_{+43.3}$ | 60.2$_{+42.9}$ | 72.3$_{+30.2}$ | 42.6 |
| SIM-FSA | 77.7$_{+53.3}$ | 78.4$_{+47.3}$ | 67.5$_{+49.9}$ | 70.0$_{+55.3}$ | 62.8$_{+46.4}$ | 69.9$_{+46.2}$ | 71.8$_{+47.5}$ | 79.4$_{+24.7}$ | 45.3 |
| STD-FSA | 79.2$_{+39.1}$ | 79.2$_{+15.9}$ | 70.6$_{+24.3}$ | 69.4$_{+11.2}$ | 67.6$_{+21.5}$ | 72.4$_{+18.6}$ | 74.8$_{+15.0}$ | 81.1$_{+7.8}$ | 18.5 |
| *Admix*-FSA | 79.4$_{+59.3}$ | 78.5$_{+39.5}$ | 68.9$_{+47.8}$ | 70.8$_{+49.4}$ | 65.0$_{+44.4}$ | 71.4$_{+42.5}$ | 70.0$_{+41.9}$ | 79.9$_{+16.4}$ | 40.2 |
| BSR-FSA | **90.1**$_{+36.3}$ | **89.4**$_{+38.8}$ | **82.8**$_{+52.0}$ | **84.2**$_{+41.9}$ | **82.2**$_{+49.2}$ | **86.6**$_{+42.8}$ | **81.6**$_{+52.8}$ | **90.9**$_{+18.0}$ | 41.5 |

### 4.4   Ensemble Models

In the field of adversarial attacks, ensemble attacks, which generate adversarial examples by leveraging gradient information from multiple source models simultaneously, have been proven effective in improving transferability. To verify the compatibility of our proposed FSA method with ensemble attacks, we generate adversarial examples using four standard trained models (Inc-v3, Inc-v4, IncRes-v2, and Res-101) and evaluate their ASR against the eight defense models mentioned earlier. Table 3 presents the ASR on ensemble models. In terms of average ASR, FSA alone achieves 87.7%, which is comparable to the SOTA BSR method and significantly outperforms traditional approaches such as DIM and SIM. Moreover, FSA demonstrates a distinct advantage against ensemble defense models. For instance, on the IncRes-v2$_{ens}$ defense model, FSA achieves an ASR of 87.1%, surpassing even the highly effective BSR and STD methods. These results validate that FSA-generated adversarial examples, crafted through frequency-space collaboration, can better adapt to the decision boundaries of ensemble models, further enhancing their transferability and robustness in attacking defense mechanisms.

**Table 3.** ASR (%) of eight models when adversarial examples are crafted using an ensemble of Inc-v3, Inc-v4, IncRes-v2, and Res-101.

| Attack | Inc-v3$_{ens3}$ | Inc-v3$_{ens4}$ | IncRes-v2$_{ens}$ | HGD | R&P | NIPS-r3 | NRP | FD | AVG. |
|---|---|---|---|---|---|---|---|---|---|
| DIM | 56.4 | 52.4 | 36.4 | 43.8 | 38.3 | 50.1 | 32.5 | 72.7 | 47.8 |
| TIM | 61.9 | 56.3 | 46.1 | 60.7 | 46.7 | 53.2 | 36.0 | 65.0 | 53.2 |
| SIM | 78.8 | 74.4 | 60.8 | 69.8 | 59.3 | 70.6 | 54.1 | 81.3 | 68.6 |
| *Admix* | 83.6 | 78.3 | 63.9 | 73.7 | 65.2 | 77.2 | 57.8 | 86.5 | 73.3 |
| SSA | 82.7 | 80.0 | 71.7 | 75.9 | 64.8 | 80.6 | **70.8** | 89.3 | 77.0 |
| BSR | **93.4** | 89.0 | 76.6 | 92.1 | 81.4 | **89.7** | 58.6 | **92.4** | 84.2 |
| STD | 89.7 | 86.8 | 84.6 | 88.6 | 85.0 | 87.4 | 72.6 | 88.3 | 85.4 |
| FSA | 89.7 | 88.8 | 87.1 | 88.9 | 86.4 | 86.9 | 84.8 | 88.9 | 87.7 |
| DIM-FSA | 90.8 | **89.6** | **87.2** | **89.1** | **86.7** | 88.4 | 84.8 | 89.4 | **88.3** |
| TIM-FSA | 87.9 | 87.8 | 85.0 | 85.6 | 83.7 | 84.8 | 83.3 | 87.0 | 85.6 |
| DI-TIM-FSA | 88.8 | 88.2 | 86.4 | 86.0 | 85.6 | 86.6 | **85.5** | 89.1 | 87.0 |

### 4.5    Ablation Study

To further investigate the performance enhancements of FSA, we conducted an ablation study and hyperparameter analysis by generating adversarial examples on Inc-v3 and evaluating them on four standard models, namely Inc-v3, Inc-v4, IncRes-v2, and Res-101, as well as three defense models, including Inc-v3$_{ens3}$, Inc-v3$_{ens4}$, and IncRes-v2$_{ens}$.

**1) The number of transformed images $N$.** As shown in Fig. 2(a), when $N = 2$, FSA outperforms BSR in transferability across three defense models, demonstrating higher efficiency and effectiveness. As $N$ increases, the attack performance progressively improves and stabilizes when $N > 20$. Therefore, we set $N = 20$ in our experiments.

**2) The value of decay factor $\beta$.** As shown in Fig. 2(b), the attack success rate generally improves as the decay factor $\beta$ increases. Specifically, when $\beta$ increases from 0.1 to 0.8, a significant rise in success rate is observed. However, beyond $\beta = 0.8$, the improvement plateaus, and further increasing $\beta$ yields only marginal benefits to attack performance. Therefore, we set $\beta = 0.8$ as the optimal hyperparameter to balance attack effectiveness and stability.

**3) The value of pyramid layers $n$.** In order to evaluate the impact of the number of pyramid layers on model robustness, an ablation study was conducted with $n = 2,3,...,9$. As illustrated in Fig. 2(c), the attack success rate remains relatively stable across different settings. Consequently, we selected $n = 5$ as a balanced configuration, which offers competitive performance while preventing potential redundancy and unnecessary computational costs introduced by deeper pyramids.

**4) The size of Gaussian kernel $K$.** To assess the impact of the Gaussian kernel size on attack performance, we conducted an ablation study varying the pyramid layer kernel from $3 \times 3$ to $11 \times 11$, as illustrated in Fig. 2(d). The attack success rate remains relatively stable across all kernel sizes, with fluctuations typically within a narrow margin (less than 2%). Specifically, the $3 \times 3$ kernel achieves comparable results to larger kernels, while maintaining computational efficiency and avoiding potential over-smoothing effects introduced by excessively large kernels. Given the marginal differences in performance and the increased computational cost associated with larger kernels, we adopt the $3 \times 3$ kernel in our final setting.

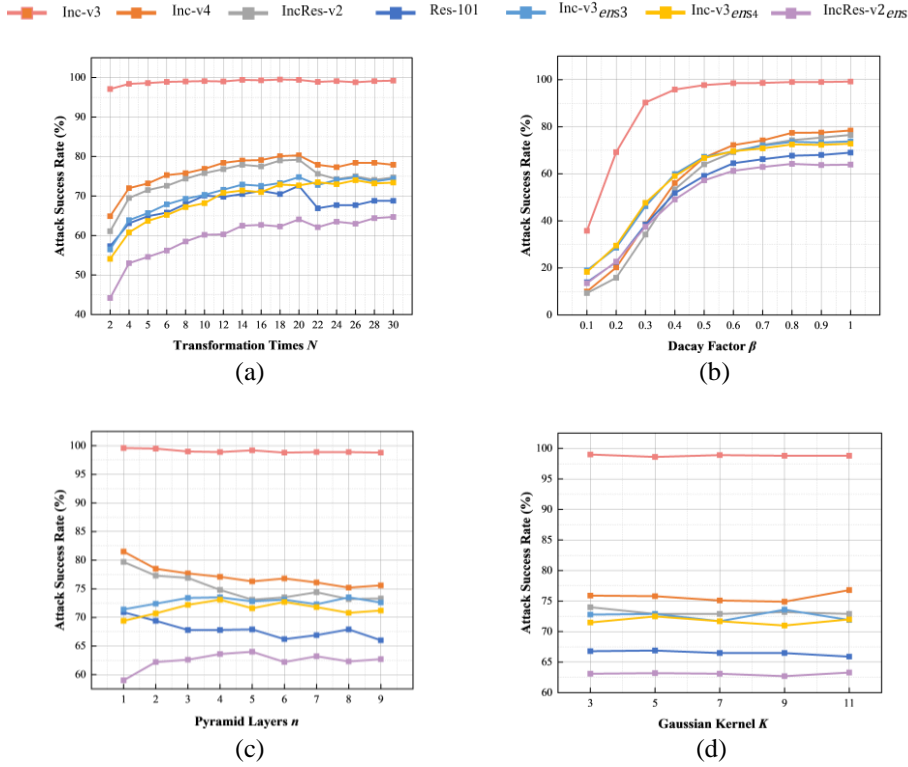

**Fig. 2.** ASR (%) of adversarial examples generated by FSA under varying numbers of images $N$, perturbation budget $\beta$, iteration steps $n$, and Gaussian kernel $K$.

**5) The values of spectral factor $\rho$ and standard deviation $\sigma$.** To ensure that the perturbation scale is consistent with the maximum perturbation magnitude, we conduct a variance ablation study using multiples of $\epsilon$. As shown in Fig. 3, when $\sigma$ remains within a lower range ($\epsilon - 3\epsilon$), the attack success rate remains high. However, when $\epsilon$ exceeds $3\epsilon$, excessive Gaussian noise may interfere with the optimal perturbation direction, leading to a decline in attack effectiveness. Additionally, as $\rho$ increases from 0.1 to 0.7, the attack success rate generally improves, but further increasing $\rho$ to 0.9 results in a slight decline. This phenomenon may be attributed to excessive spectral enhancement, which disrupts the noise distribution and reduces the attack's generalization capability. Based on the experimental results, the combination of $\sigma = 2\epsilon$ and $\rho =$ 0.7 consistently achieves high attack success rates across most target models, making it the optimal configuration for our final experimental setup.

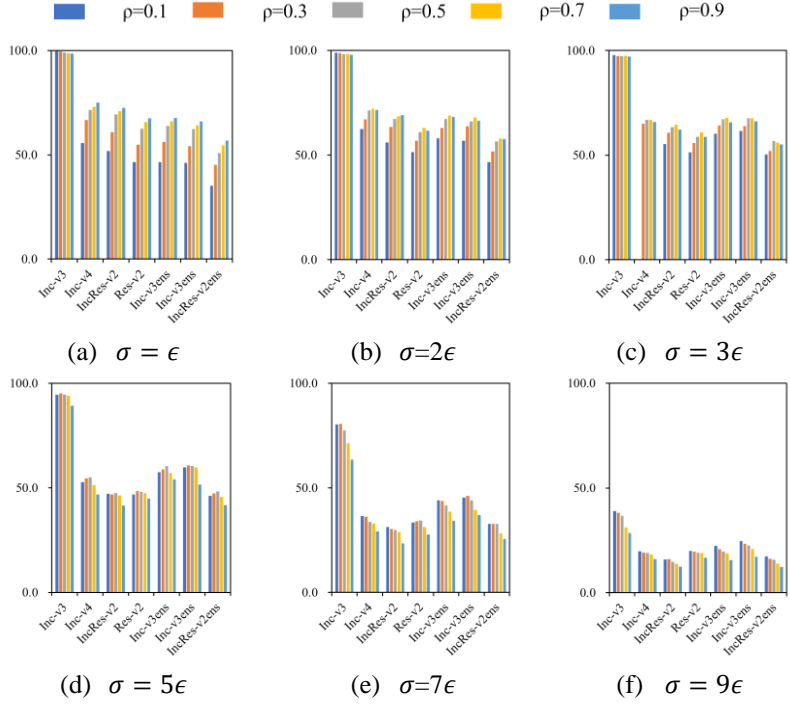

**Fig. 3.** ASR (%) of Adversarial Examples Generated by FSA with Different Values of spectral factor $\rho$ and Standard Deviation $\sigma$.

**6) The Contribution of HAM and HFM**. In order to comprehensively evaluate the standalone effectiveness of each component within the FSA framework, a detailed module-level ablation study was conducted. As illustrated in Fig. 4, both modules demonstrate a marked improvement in performance when applied independently, surpassing the baseline method (MI-FGSM) across all target models. This finding indicates that each component contributes significantly to the generation of adversarial examples. Specifically, HAM significantly enhances attack transferability by introducing frequency-based augmentations to increase input diversity, while HFM improves attack robustness and stability through multi-scale gradient decomposition and fusion. While each module demonstrates strong performance in isolation, the integrated FSA configuration, which incorporates both HAM and HFM, attains the highest overall performance, surpassing all standalone configurations. These results validate the complementary roles of spatial- and frequency-domain manipulations and confirm the effectiveness and rationality of FSA's modular design.

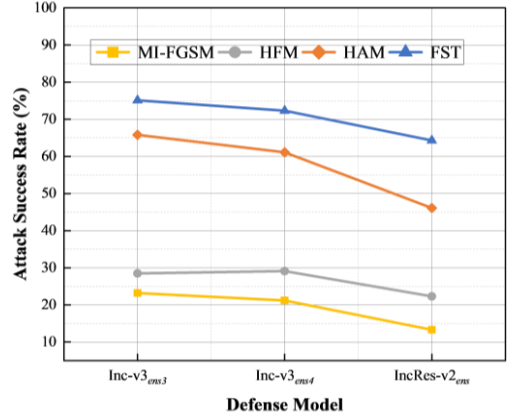

**Fig. 4.** ASR (%) of FSA modules across three target models (Inc-v3$_{ens3}$, Inc-v3$_{ens4}$, Inc-Res-v2$_{ens}$). MI-FGSM serves as the baseline.

### 4.6    Visualization of Adversarial Examples

We randomly select several adversarial examples for visualization, which are obtained by utilizing MI, BSR and FSA to attack Inc-v3. As shown in Fig. 5, five pairs of clean images and their corresponding adversarial counterparts are presented. As observed, the perturbation intensity introduced by our method is comparable to that of baseline methods, with no noticeable difference.

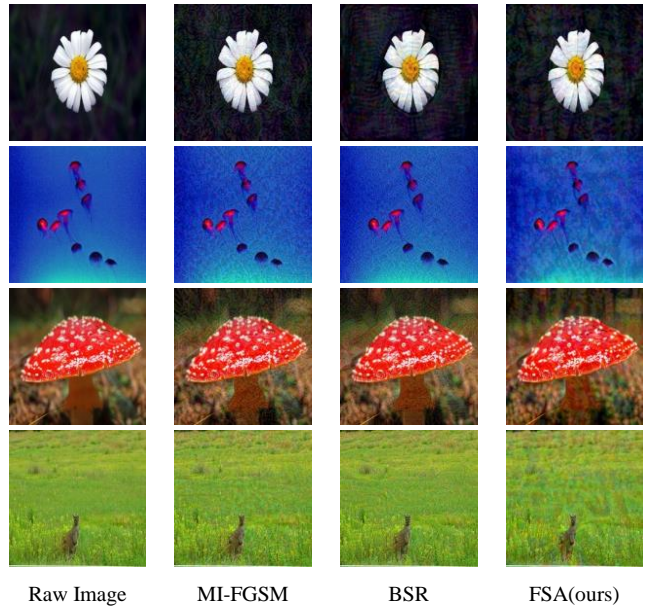

Raw Image        MI-FGSM        BSR        FSA(ours)

**Fig. 5.** Visualization of randomly selected raw images and their corresponding adversarial examples (which are crafted on Inc-v3).

## 5     Conclusion

In this paper, we propose a new Frequency-Space Attack (FSA) framework that combines frequency-domain and spatial-domain transformations to enhance the effectiveness of adversarial attacks, particularly against black-box defense mechanisms. By utilizing High-Frequency Augmentation and Hierarchical-Gradient Fusion, FSA significantly boosts the transferability of adversarial attacks by emphasizing high-frequency components and capturing both global and fine-grained features. Our experimental results show that FSA outperforms current state-of-the-art methods. These findings highlight the potential of combining frequency-domain techniques with traditional spatial-domain approaches to develop more robust and efficient adversarial attacks, advancing the field of machine learning security.

**Acknowledgments.** This research is funded by the Scientific Research Foundation of Shantou University under Grant NTF24001T.

**Disclosure of Interests.** The authors declare that they have no known competing financial interests or personal relationships that could have appeared to influence the work reported in this paper.